# AVOID: Autonomous Vehicle Operation Incident Dataset Across the Globe

**Authors**

Ou Zheng, Mohamed Abdel-Aty, Zijin Wang*, Shengxuan Ding, Dongdong Wang, Yuxuan Huang

**Affiliations**

1. Department of Civil, Environmental and Construction Engineering, University of Central Florida, Orlando, FL 32816, USA

Corresponding author(s): Zijin Wang (zijinwang@Knights.ucf.edu)

## Abstract

Crash data of autonomous vehicles (AV) or vehicles equipped with advanced driver assistance systems (ADAS) are the key information to understand the crash nature and to enhance the automation systems. However, most of the existing crash data sources are either limited by the sample size or suffer from missing or unverified data. To contribute to the AV safety research community, we introduce AVOID: an open AV crash dataset. Three types of vehicles are considered: Advanced Driving System (ADS) vehicles, Advanced Driver Assistance Systems (ADAS) vehicles, and low-speed autonomous shuttles. The crash data are collected from the National Highway Traffic Safety Administration (NHTSA), California Department of Motor Vehicles (CA DMV) and incident news worldwide, and the data are manually verified and summarized in ready-to-use format. In addition, land use, weather, and geometry information are also provided. The dataset is expected to accelerate the research on AV crash analysis and potential risk identification by providing the research community with data of rich samples, diverse data sources, clear data structure, and high data quality.

| Measurement(s) | Automated Driving System, Advanced driver Assistance System, Crashes |
|---|---|
| Technology Type(s) | Large Language Model, Google Maps, OpenStreetMap, Manual Entry, OpenweatherAPI |

| Factor Type(s) | Temporal Interval |
|---|---|
| Sample Characteristic-Location | United States, China, Australia, Argentina, Spain, Portugal, France, United Kingdom, Ireland, France, Belgium, Italy, Albania, Greece, Ukraine, Russia, India, New Zealand |

Machine-accessible metadata file describing the reported data: https://github.com/ozheng1993/AVOID-Autonomous-Vehicle-Operation-Incident-Dataset

# Background & Summary

Autonomous vehicles (AV) have become a reality in recent years, and as this technology continues to advance, it is important to ensure that these vehicles are safe and reliable. With the potential to greatly reduce the number of accidents caused by human error, AVs are seen as a promising solution for improving road safety. According to the National Highway Traffic Safety Administration (NHTSA), 94% of all traffic accidents are caused or related to human error [1]. By removing human error from the equation, AVs could significantly reduce the number of accidents on the roads. Additionally, AVs are equipped with advanced safety features such as collision avoidance systems and advanced sensors, which can help to prevent accidents. These safety features are constantly monitoring the vehicle's surroundings and can react much faster than a human driver in the event of an unexpected situation.

However, crashes involving autonomous vehicles do occur, and it implies that much effort is still needed to enhance AV safety. Much research has been conducted to investigate AV safety issues from the perspectives of public perception, crash pattern, crash reasoning, pre-crash scenarios, crash severity, liability determination, and crash avoidance. The public's attitudes towards AVs play an important role in their development and adoption, and it is reported that significant negative impact was observed after the occurrence of an AV fatal crash [2]. Also, the study by Kalra and Paddock claimed that hundreds of millions of miles and sometimes hundreds of billions of miles are needed to demonstrate AV reliability in terms of fatalities and injuries [3]. Learning from AV crashes is an effective approach to understanding their crash nature and identifying the potential risk. It has been found that intersection-related features are the key predictors of Vulnerable Road Users (VRUs)-AV-related crashes [4]. Liu et al. compared the pre-crash scenario typology between conventional vehicles and AVs and indicated that in 52.46% of the AV crashes they are rear-ended by conventional vehicles [5]. Many work focused on modeling the crash severity and crash type contributing factors [6-9]. Also, autonomous driving disengagement issue was investigated by Favaro et al [10]. With this knowledge absorbed, collision avoidance algorithms are developed for AVs that work on various conditions [11-13] . Furthermore, the challenges and approaches of realizing AV safety are discussed [14].

The value of AV operational data has been validated in the above-mentioned research work, and in particular, the AV crash data provides crucial information about AV safety issues. However, due to the small amount of test AVs on the road, the sample size of AV crashes is still at a small scale with imbalanced and underreported issues, making it difficult to conduct related analyses. Although there are a few studies that summarize the AV crashes from different data sources, these datasets only provide limited records [15]. Apart from the crash data, some other data sources provide statistics about AV operational and safety circumstances. In Table 1 we identify and summarize the available data sources related to AV operation and safety.

**Table 1. Summary of available data sources**

| Data type | Description | Data link |
|-----------|-------------|-----------|
| AV crash dataset | Open-source AV crash and disengagement dataset for crashes from 2014 to 2020, and supplementary road network and land-use data extracted from OpenStreetMap | https://figshare.com/articles/dataset/Autonomous_Vehicles_Crash_and_Disengagement_Report_CA_DMV_/16923463/1 |
| NHTSA AV crash data archive | AV and low-level automation vehicles crash data since July 2021 that are summarized in csv files. | https://www.nhtsa.gov/laws-regulations/standing-general-order-crash-reporting#data |
| California DMV AV crash report | Raw crash report from California DMV, which requires all AV crashes to be reported and opened. Masks in the report may be encountered. | https://www.dmv.ca.gov/portal/vehicle-industry-services/autonomous-vehicles/autonomous-vehicle-collision-reports/ |
| NHTSA AV general information catalog | Official documents and resources on AV policies, general order, testing guidelines and records, education materials, etc. | https://www.nhtsa.gov/technology-innovation/automated-vehicles-safety#resources |
| NHTSA AV operation resource | Web pilot that provides the public with AV testing information, including testing vehicles search and tracking. | https://avtest.nhtsa.dot.gov/av-test |

This paper aims at introducing Autonomous Vehicle Operation Incident Dataset (AVOID): an open-source dataset that provides information on AV crashes. Compared with existing datasets, AVOID contains much more rich information in terms of sample size, vehicle categories, and data sources. The vehicles included in the dataset are divided into three classes: ADAS vehicles (vehicles equipped with advanced driver assistance system, corresponds to SAE automation level 2), ADS vehicles (vehicles that are capable of operating fully autonomously, corresponds to SAE automation level 3-5), and low speed shuttles (corresponds to SAE automation level 3-5). The dataset fuses multiple data sources, including more than 1000 crash reports, traffic data, environmental data, and news reports. Pre-crash conditions, event nature, and road and environmental features are collected and summarized as the main components of the dataset. By opening this dataset, we aim to accelerate the research on studying and understanding the AV crash nature, and to identify ways to improve the safety of the AVs.

# Methods

The dataset generation consists of three work stages: raw data collection, data processing and integration, and manual checking. The raw data are collected from the National Highway Traffic Safety Administration (NHTSA), California Department of Motor Vehicles (CA DMV), worldwide news and social media, Google Maps, OpenstreetMaps (OSM), and Open Weather API. However, these data sources are with different formats and are even cross-modal, and automatic and manual processing methods are combined to organize and integrate the data. As missing/duplicate or unverified records may appear in the raw data, manual checking was applied. The workflow of the dataset generation is shown in Figure 1.

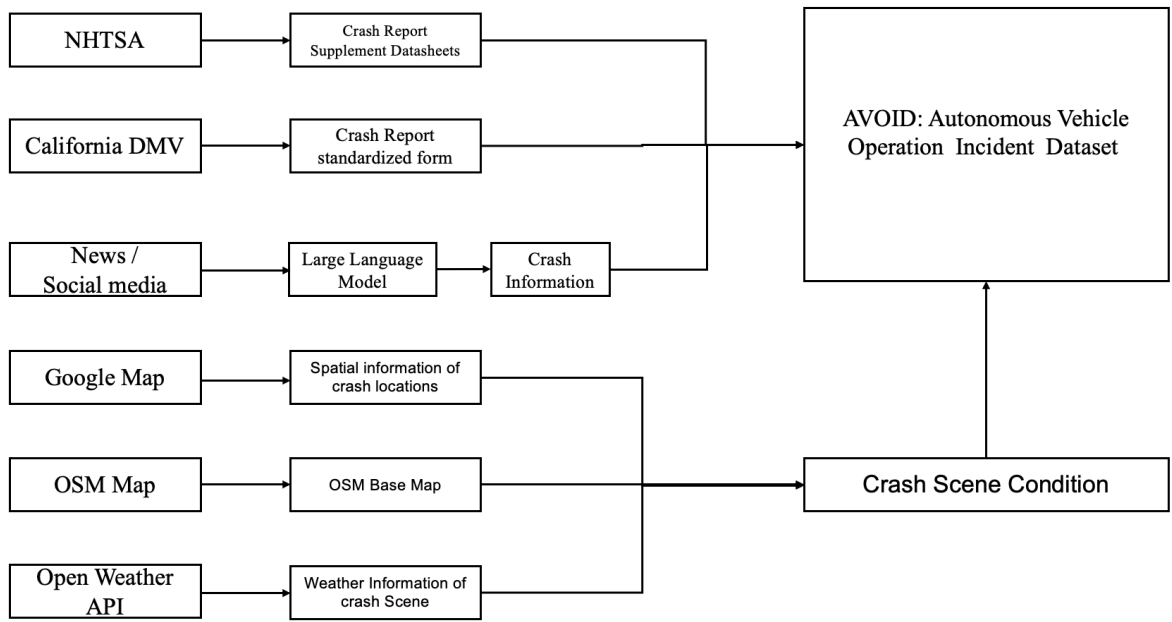

**Figure 1. Workflow of dataset generation**

# Official Crash Report

## NHTSA

The National Highway Traffic Safety Administration (NHTSA) has implemented a Standing General Order that mandates manufacturers and operators to report specific crashes involving vehicles that are equipped with automated driving systems or advanced driver assistance systems. The purpose of this General Order is to ensure that NHTSA receives immediate and transparent notification of actual crashes related to these types of vehicles from manufacturers and operators. The collected data would enable NHTSA to gain insights into real-world crashes associated with ADS and SAE Level 2 ADAS vehicles. Under the Standing General Order issued by the National Highway Traffic Safety Administration, entities identified as manufacturers or operators must report any crash that involved the use of automated driving systems (ADS) within 30 seconds of the crash and resulted in property damage or injury [16]. Similarly, if the crash involved a Level 2 advanced driver assistance system (ADAS) and resulted in a fatality, a vehicle tow-away, an air bag deployment, or any individual being transported to a hospital for medical treatment, or if the crash involved a vulnerable road user, the entity must report it within the same 30-second time frame. Crash information can be obtained from multiple sources, such as the NHTSA Crash Investigation Sampling System, Crash Report Supplement Datasheets, and Special Crash Investigations.

## California Department of Motor Vehicles

CA DMV requires all manufacturers who are testing autonomous vehicles to report any collision that resulted in property damage, bodily injury, or death within 10 days of the incident using the OL316 standardized form. AV crash reports from 2014 to 2023 are currently available. Those crash reports provided various variables such as pre-crash conditions and crash scene information. Specifically, the pre-crash condition variables considered were vehicle manufacturers, AV driving modes, pre-crash vehicle movements status, third-party status. On the other hand, the crash scene information included the weather condition, lighting condition, type of collision, crash narratives from the reporting agency, etc. The information is extracted with our self-developed PDF reader and summarized in format aligned with NHTSA data sheets. Manual checking is conducted to correct the detection or missing values.

# Auxiliary Crash Information

Since AVs just recently emerged in the last few years, the official records of AV crashes are still limited, which may hinder larger scale data analysis research. To tackle this potential challenge, we prepared auxiliary crash information by integrating other data sources, including news and posts from social media that reports AV crashes with crash description, crash video, and driver narratives. Social media posts also enable sentiment analysis which is important to understand the public's concerns and acceptance of the technology. As the social media and digital news data volume is extremely large, the text mining techniques plays an important role in the data collection and processing. To efficiently extract information, we incorporate large language models (LLMs) like ChatGPT into the processing pipelines. With LLMs, a variety of information is extracted with prompt questions.

## Worldwide Mass Media Data

Although the United States is the only country that provides a comprehensive AV-related crash reporting system so far, some other countries, including Australia, Canada, China, Germany, New Zealand, and the United Kingdom, have also initiated government-level discussion on AV operation and regulation [17], which are widely reported by mass media. Also, AV safety always draws great attention of the public, resulting in some AV crashes being reported and reposted through news agencies or the general public on social media. Fifty two crash news is captured in total, and we will keep updating the dataset once new samples are obtained. Given different formats of mass media data, we conduct different curations to prepare auxiliary datasets.

## Textual Data Curation

Collecting crash information from different countries can be challenging due to the language barrier. Language differences can make it difficult to accurately collect and interpret information, especially if there are nuances or idioms that are unique to a particular language or culture. It is also expensive to have a multilingual staff or translators who can help bridge these communication gaps and ensure that the information is correctly collected and analyzed. Therefore, we adopt machine translation to overcome data limitations due to different languages. The crash news from non-English speaking countries is translated with language models for text analysis. Figure 2 shows a German news report about an AV crash carried out with ChatGPT and the final extracted information is unified into English.

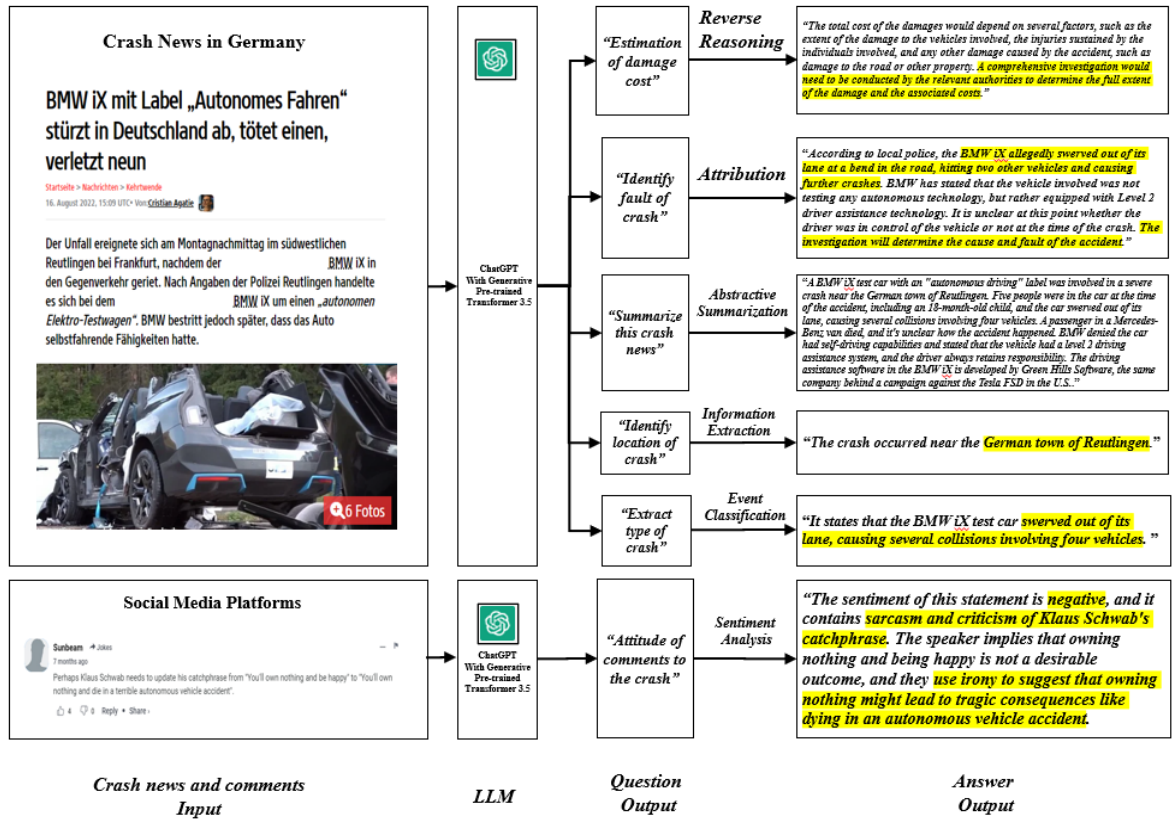

**Figure 2. Example of Textual Data Curation**

## Video/Audio Data Curation

Next, news or social media videos pertaining to crashes are collected for information extraction. Figure 3 shows the pipeline of how to extract crash-related information from the news report. By employing vision-language modeling to understand video scenes, thus converting the videos from social media or digital news into textual data. The vision-language modeling is implemented with the assistance of LLMs and large Image-to-Text models like BLIP. The frame images are fed into BLIP to generate image captions and these captions derive prompt questions to ChatGPT for crash information generation. The generated texts are further extracted and formatted for final crash report preparation. In addition, audio from media is gathered to enrich the dataset. We conduct large Speech-to-Text models to convert verbal narratives from the news or social media to textual descriptions. With textual data, text mining is further conducted to generate key information for crash reports. In addition, the crash scenes are also reconstructed with provided keywords.

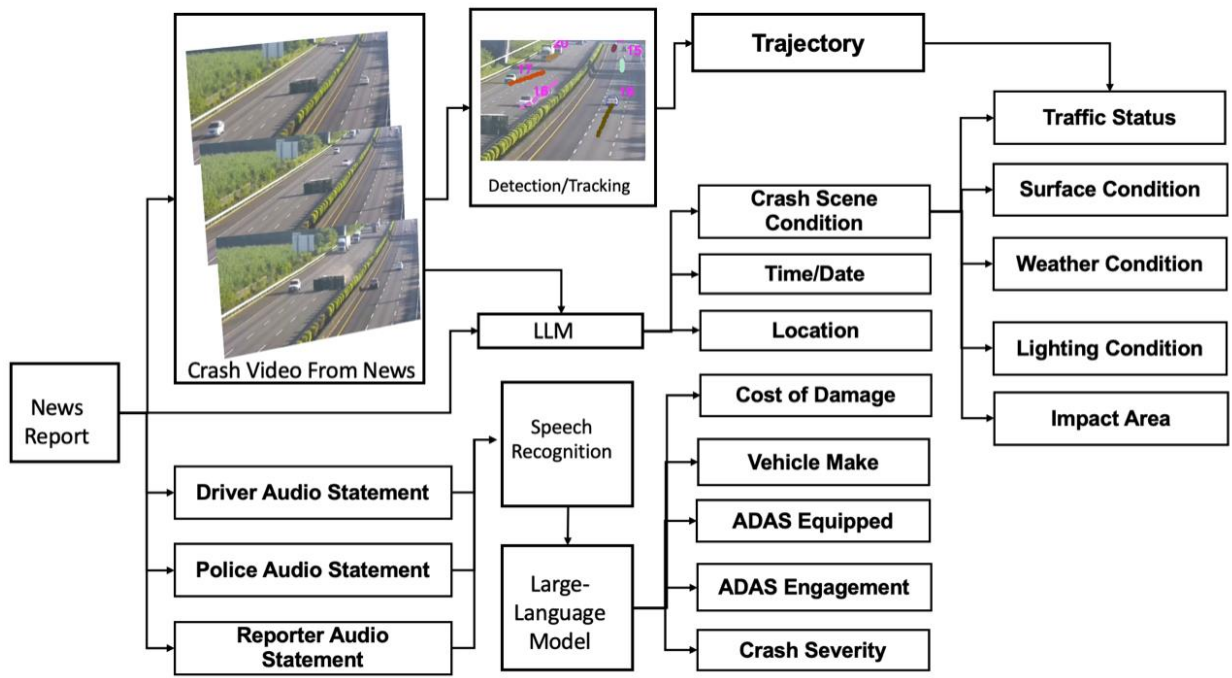

**Figure 3. Pipeline of crash data curation from news**

Apart from the conventional crash elements that could be obtained through crash reports, the news and social media posts provide extra information about the crash as the crash video, driver narratives, and public perception are available. As Figure 3 shows, multi-modal information can be extracted through LLM, including the traffic status, surface condition, weather condition, lightning condition, and the vehicle trajectory, etc. Benefiting from this, detailed analyses can be conducted at a more microscopic level by reconstructing the trajectories from crash videos or modeling the vehicles' pre-crash maneuvers through driver narratives. For instance, the extracted trajectory, as shown in Figure 4, can be used to identify the vehicle's pre-crash kinematic status including braking and steering behavior.

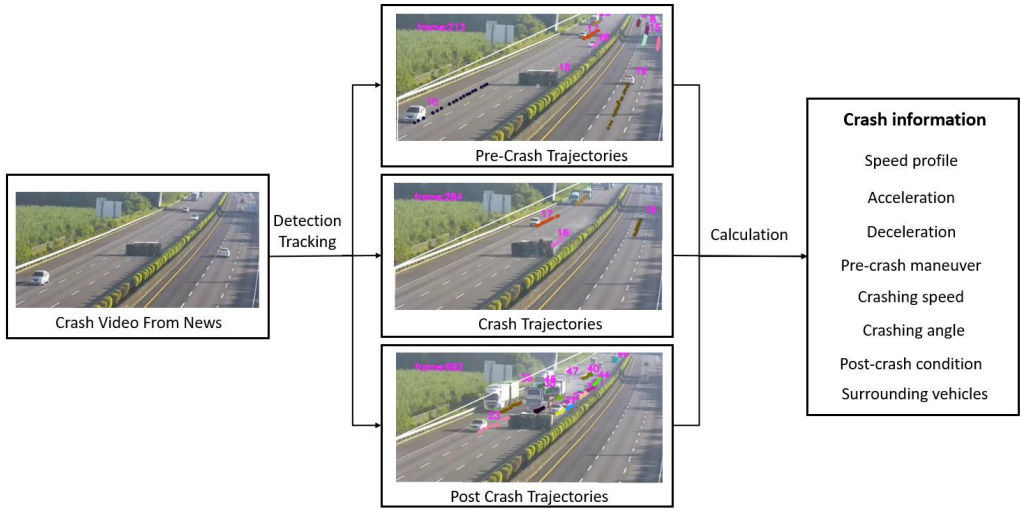

**Figure 4. Trajectory extraction from video and its usage**

# Crash Scene Condition

## *Spatial information of crash locations*

To examine the features of the location where each AV-related accident happened, OpenStreetMap (OSM) was used to identify the location of every crash and extract variables related to land use, building, vegetation, and road segment types. As Figure 5 shows a custom python code based on GeoPandas and OSMnx was utilized to extract information from OSM on land use, vegetation, and other spatial information of crash locations. Moreover, a Google Maps satellite image was also obtained through the Google Maps API to validate the output from OSMnx.

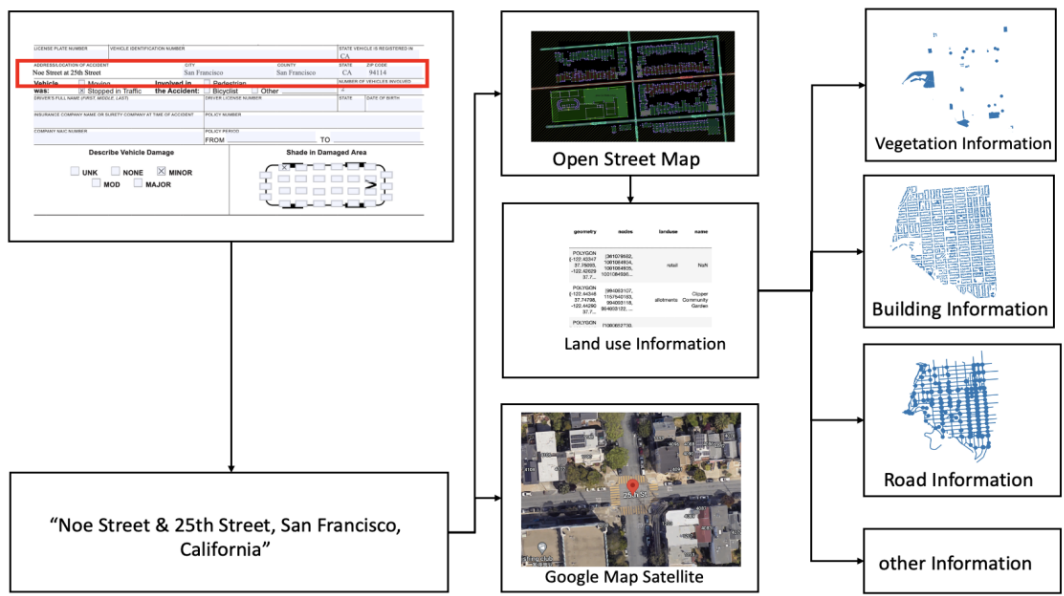

**Figure 5. Pipeline of extra Spatial information of crash locations**

*Weather*

The OpenWeather API is a weather data API that provides access to real-time, historical, and forecast weather data for various locations around the world. We can request it by the following URL "*http://api.openweathermap.org/data/2.5/onecall/timemachine?lat={latitude}&lon={longitude}&dt={unix timestamp}&appid={your api key}*".The latitude and longitude coordinates of the location of interest should be included by replacing the placeholder values {latitude} and {longitude} with the appropriate values. Additionally, a Unix timestamp for the date of interest must be included, which represents the number of seconds since January 1, 1970. This timestamp should replace the placeholder value {unix timestamp}. Lastly, the user must include his/her own API key in the request by replacing {your api key} with his/her own unique key. The API will return a JSON response containing the historical weather data for the specified location and date.

# Data Record

The processed dataset comprises crash data from 2014 to March 12th, 2023, which was reported by the California Department of Motor Vehicles (CA DMV) and the National Highway Traffic Safety Administration (NHTSA) and regularly updated. Moreover, additional crash data were obtained from news or social media sources. Table 2 demonstrates the dataset structure. The data consists of three main components: the processed and integrated data (AVOID), and the raw data collected from various data sources, and the code for extracting weather and land use information.

## AVOID

The AVOID dataset contains four comma-separated values (CSV) files: "ADS (SAE L3-5).csv", "ADAS (SAE L2).csv", "Shuttle (SAE L3-5).csv", and "Crash_news(SAE L2).csv", which are extracted and organized from CA DMV reports (since 2014), NHSTA data sheets (since July 2021), and news reports. The three csv files are generated by merging the NHTSA data and CA DMV reports while filtering out one of the overlapped records, and the three files are organized with the same format. The crash features are divided into seven general categories: vehicle conditions, pre-crash conditions, road and environment, crash outcome, contact areas (ego vehicle), contact areas (the other vehicle). The summary of the crash features along with their definitions are attached in the appendix. Figure 6 shows the AVOID's basic statistics and Figure 7 shows the crash features distributions (excluding news reports data due to missing values). It is observed from Figure 6 (a) that crashes most frequently occurred in the afternoon peak hours. Figure 6 (c) demonstrates that 58%, 25%, and 17% of the crashed vehicles are ADAS-engaged, ADS-engaged, and disengaged when the crash happened, respectively. Figure 6 (d) shows the trend of the crash number by year, and a spike in 2021 is observed because the ADAS vehicle crash data are collected since then. The lighting condition distribution is shown in Figure 7 (a), indicating most of the crashes had happened under daylight. Figure 7 (b) shows the frequency of the collision spot on the autonomous vehicles (including ADAS and ADS). It could be observed that the most impacted area is the front of the vehicles, which includes the AV rear ending, hitting another vehicle from the side or hitting a fixed object. This is a common problem with AVs not recognizing or confusing a hazard with the background. It needs to be mentioned that a crash may have multiple impact areas on the vehicle. The crash severity distribution is shown in Figure 7 (c), and it should be noticed that a large portion of the crash reports do not provide this information. The pre-crash maneuver distribution is shown in Figure 7 (d). Figure 7 (e) demonstrates that apart from the missing values, most of crashes appeared on highway or freeway and intersections. Lastly, most of the crashes occurred at a relatively low speed, as Figure 7 (f) shows.

The aggregated data sheets for news reports (Crash_news (SAE L2).csv) summarizes key elements of the crashes, including time, location, weather, roadway type, collided objects, impact area, crash severity, and most importantly the drivers' narratives. As the news are prone to report severe accidents, the crashes we collected have a much higher injury severity level compared to the conventional crash database, with 43.8% of the crashes having a fatality or an incapacitating injury. Also, 47.3% of the crashes are for vehicles hitting an object (e.g., wall, mid-block). It should be noted that missing values exist due to the limited information obtained from the news reports. Figure 8 illustrates the crash locations on the world map, indicating most of the crashes are from the US and a small portion are from China, Europe, and other regions, which are obtained from NHTSA and global wide news reports.

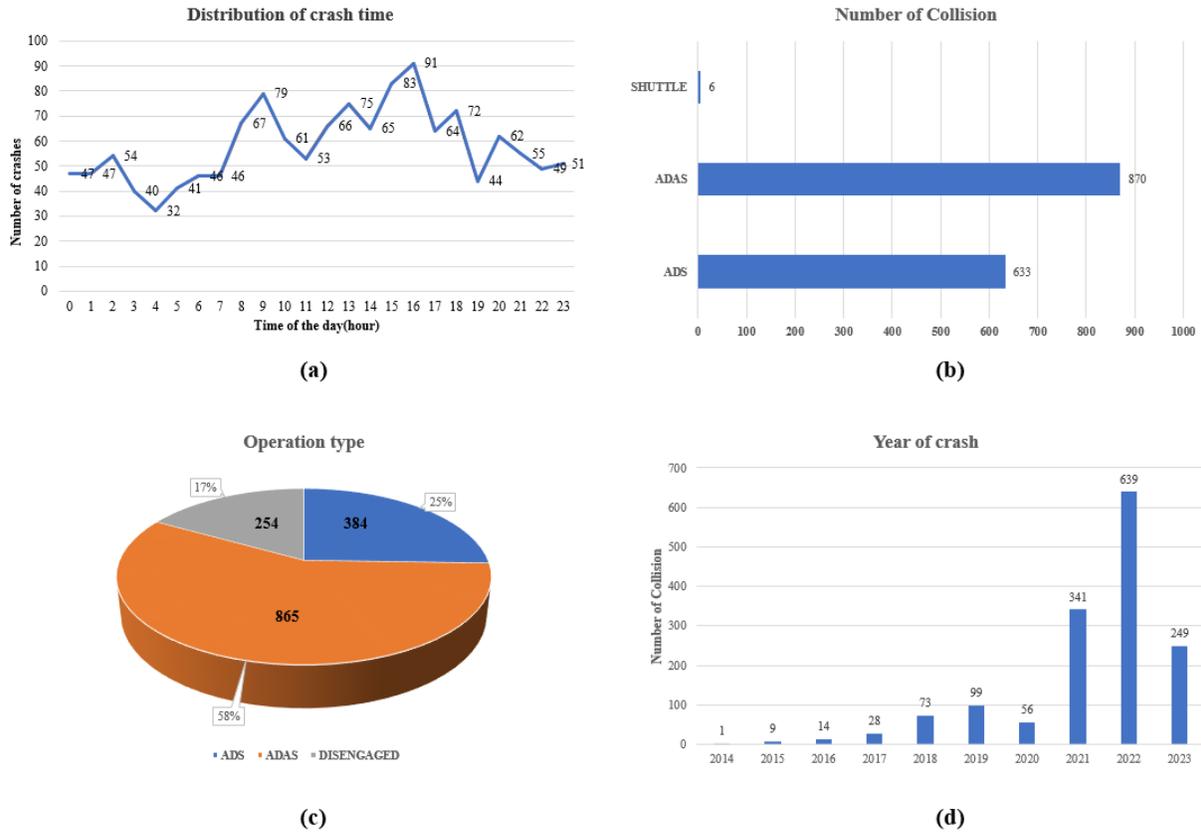

**Figure 6. Visualization of basic dataset features.**

(a) Hourly distribution; (b) number of crashes by the three defined vehicle types; (c) proportion of automated system engagement types; (d) crash frequency by year

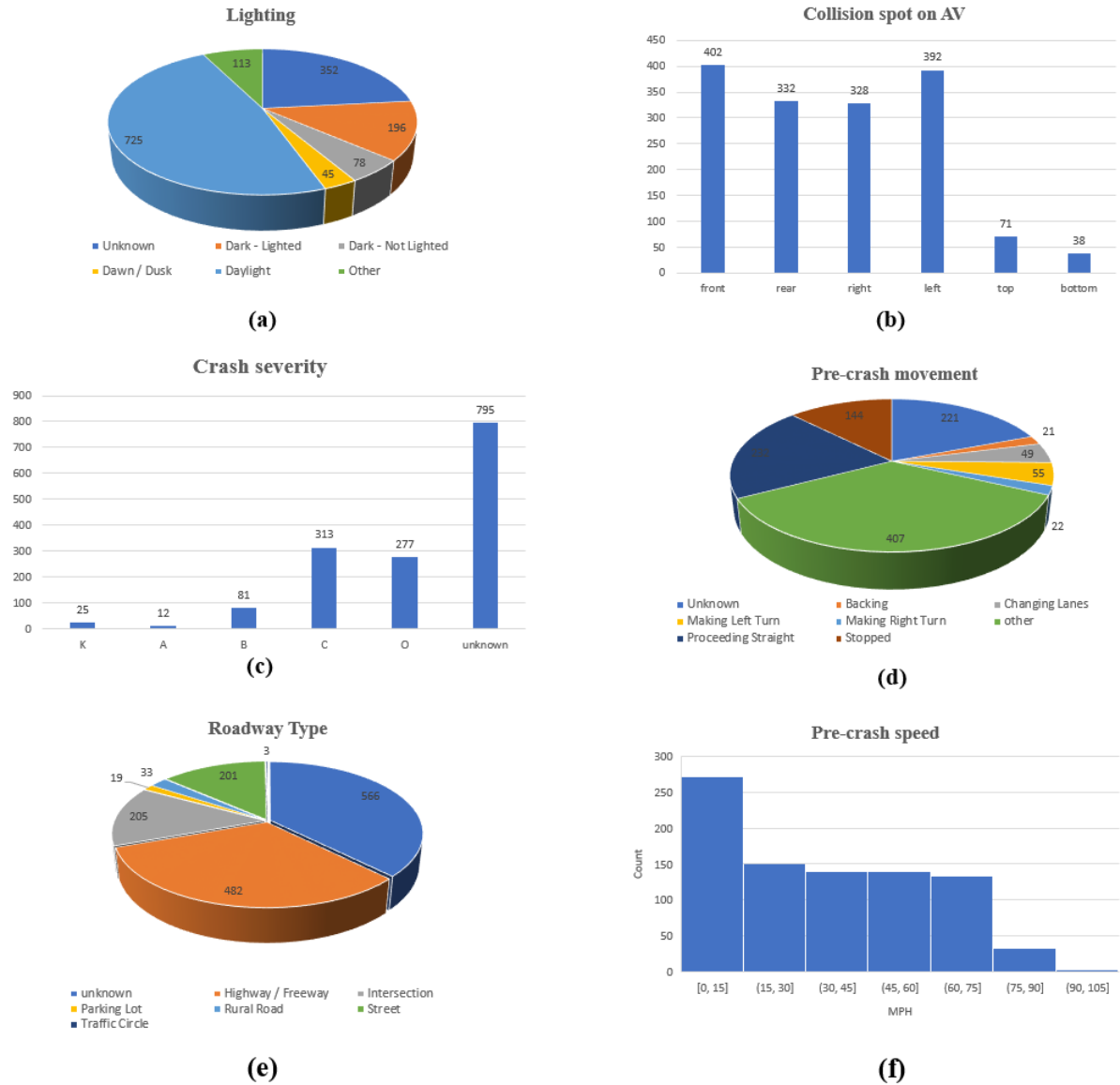

**Figure 7. Visualization of crash features.**

(a) Lightning distribution; (b) frequency of collision spot of autonomous vehicles (a crash may have multiple impact areas); (c) crash type frequency; (d) pre-crash movement distribution; (e) crash roadway type distribution; (f) histogram of pre-crash speed

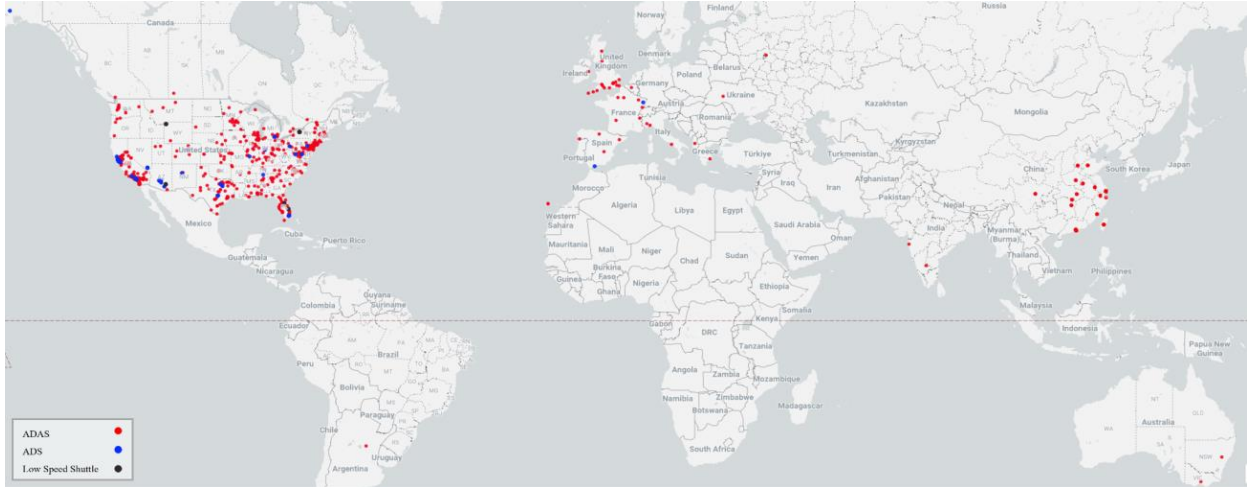

**Figure 8. Plot of Crash Location on World Map.**

# Raw Data

The raw data is also stored in the dataset webpage (under the *Data>RAWData* folder as shown in Table 2), including CA DMV crash reports and disengagement reports, NHTSA crash report summary, and news/social media data. The DMV reports are in PDF format while the NHTSA sheets are in csv format. In total, 560 crash reports and 687 disengagement reports were downloaded from CA DMV, and the downloaded NHSTA data sheets contain 3844 records (duplicate records exists). The NHTSA also provides a small portion of data outside of the US. For the news reports, 52 records are currently captured and the link to the original news reports are stored in the dataset webpage. The news reports data will be expanded regularly in the following work. All the raw data are stored in Google Drive space with a link on the dataset webpage.

**Table 2. Dataset structure**

| Directories | | | | Description |
|---|---|---|---|---|
| /Data | /AVOID | | | Processed Crash Data |
| | /RAWData | /NHTSA | | Crash report in csv format collected from from 2021-2022 |
| | | /CA DMV | /CollisionReport | Collision report in PDF format collected from CA DMV from 2014 to 2023 |
| | | | /DisengagementReport | Disengagement report in PDF format collected from CA DMV from 2015 to 2023 |
| | | /NewsMedia | | ADS or ADAS new report/video/audio collect from worldwide |
| | /Code | | | Code from process spatial information |

# Technical Validation

## Crash Description Validation

The important sources of crash data are NHTSA and California DMV, which are validated by the officials and further prepared with sanity checks. For the sanity check, the first step is to verify the incomplete data records. For example, the reports from NHTSA show the absence of certain records for incidents. Before filling these records, manual checking is conducted to verify the blanks and ensure the correctness of existing partial records. The second step for the sanity check is to delete duplicate records. The validation is performed by geospatial and temporal record checking. Due to the duplication of fields such as "VIN" in the data, the records with the same location and time information are viewed as duplicated ones and duplicated records are removed. After flittering the data based on the spatiotemporal relationship, the data is checked according to the fields such as "Vehicle ID", "Report ID" and "Same Vehicle ID".

## News/Social Media Data Validation

For unofficial raw data like news and social media postings, the information is collected from reliable publishing platforms, including government official news or mainstream news. The contents are reformatted through text processing with Python scripts and validated with sanity checks manually to ensure text quality.  To generate crash reports from this information, the prompt questions for LLM like ChatGPT are optimized with the voting across multiple specialists in crash report documentation. The generated results are further evaluated by crash report experts to ensure accuracy, completion, and reliability.

## Geospatial Data Validation

In addition to the coordinates provided in official crash reports, Geospatial data are collected from OpenStreetMap and cross-validated with GoogleMap. The location is determined with direct or generated geospatial description. The direct geospatial description is obtained from reports, news, or social media postings through text mining. The locations generated from visual data are validated with manual verification of key landmarks in the crash scenes from Google Maps satellite images and street view.

## Weather Data Validation

Weather data are downloaded from Open Weather API and reformatted by Python scripts. Based on the coordinates and the crash time, Open Weather API is used to impute missing data and add field of visibility. The historical records are validated manually to ensure the correct time and locations. The correctness of multi-source data fusion is verified by cross-referencing with the meteorological conditions in the current tables.

# Usage Notes

The processed data is stored in CSV files, which can be opened using spreadsheet software like Microsoft Excel or analyzed using programming languages such as Python or R.

# Code Availability

The codes for data validation and processing are available in the published GitHub repository. https://github.com/ozheng1993/AVOID-Autonomous-Vehicle-Operation-Incident-Dataset

The quick tutorial and README file are also included in the repository for reference. Python scripts for geospatial data processing are prepared with the OSMnX package and offered in the repository, which can be referred to in the file Address2OSM.ipynb under the folder of code. Any updates will be published on GitHub.

# Acknowledgments

This paper is based on "Autonomous Vehicles Collision Reports" released by the California Department of Motor Vehicles from 2014 to 2023 and NHTSA's "Standing General Order on Crash Reporting" supplyment datasheet.

# Contributions

Ou Zheng: Conceptualization, Data Collection, Data Processing, Software, Writing and Editing, Revision, Resources. Mohamed Abdel-Aty: Conceptualization, Resources, Data Collection, Data Processing, Supervision, Writing and Editing, Reviewing, Revision. Zijin Wang: Conceptualization, Data Collection, Data Processing, Software, Writing and Editing, Revision. Shengxuan Ding: Conceptualization, Data Collection, Data Processing, Software, Writing and Editing, Revision. Dongdong Wang: Conceptualization, Software, Resources, Reviewing. All authors have read and agreed to this version of the manuscript. Yuxuan Huang: Conceptualization, Resources, Reviewing.

# Ethics declarations

## Competing interests

The authors declare that they have no known competing financial interests or personal relationships which have, or could be perceived to have, influenced the work reported in this article.

# Appendix

| General Category | Variable | Definition | Data Type |
|---|---|---|---|
| Vehicle conditions | Reporting Entity | The full name of the Reporting Entity filing the report. | Text (128) |
| | Make | The make of the subject vehicle as branded, as reported by the Reporting Entity. | Text (25) |
| | Model | The model of the subject vehicle as branded, as reported by the Reporting Entity. | Text (256) |
| | Model year | The model year of the subject vehicle, as reported by the Reporting Entity. | Number (4) |
| | Mileage | The odometer reading of the subject vehicle in miles at the time of the incident, as reported by the Reporting Entity. | Number (9) |
| Pre-crash conditions | Driver / Operator Type | The Reporting Entity's report of the individual responsible for the operation, fallback operation, or any part of the dynamic driving task (DDT) for the subject vehicle at the time of the incident. | Text (80)<br><br>Possible values: consumer, in-vehicle, remote, in-vehicle and remote |
| | Automation System Engaged | The Reporting Entity's report of the highest-level driving automation system engaged at any time during the period 30 seconds immediately prior to the commencement of the crash through the conclusion of the crash. | Text (48)<br><br>Possible values: ADAS, ADS, "Unknown, see Narrative." |
| | Pre-Crash Movement | The Reporting Entity's report of the pre-crash movement of any crash partner / other vehicle, or non-motorist (NM) prior to the incident. | TEXT(80) |
| | ADAS/ADS Version | The Reporting Entity's report of the hardware and software version numbers of the ADAS or ADS in use on the subject vehicle at the time of the incident. | Text (48) |
| | ADS Equipped | The Reporting Entity's report of whether the subject vehicle equipped with an ADS. | Text (48) |
| Road and environments | Roadway Type | The type of road on which the subject vehicle was operating at the time of the incident, as reported by the Reporting Entity. | TEXT(80)<br><br>Possible values: highway, intersection, parking lot, traffic circle, rural road, unpaved road |
| | Roadway Surface | The roadway surface environmental conditions at the time of the incident, as reported by the Reporting Entity. | TEXT(80)<br><br>Possible values:<br><br>Dry, snow, wet, other |
| | Roadway Description | The roadway conditions, excluding weather and surface conditions, at the time of the incident, as reported by the Reporting Entity. | TEXT(80)<br><br>Possible values: no unusual conditions, traffic incident, work zone, missing/degraded markings |
| | Speed Limit | The posted speed limit in MPH on the roadway where the incident occurred in miles per hour, as reported by the Reporting Entity. | NUMBER(5) |
| | Lighting | The lighting conditions at the time and location of the incident, as reported by the Reporting Entity. | TEXT(80)<br><br>Possible values: daylight, dawn/suck, dark-lighted, dark-not lighted, other |
| | Weather | The weather or environmental conditions at the time and location of the incident. | 9 independent variables: clear, snow, cloudy, fog/smoke, rain, severe wind, unknown, other, other text |
| Crash outcomes<br><br>Crash attributes | Was Vehicle Towed | The Reporting Entity's report of whether the Crash Partner / Other vehicle was towed from the scene of the incident. | TEXT(80) |
| | Crash With | The Reporting Entity's categorical description of any vehicle, non-motorist, animal, or object with which the subject vehicle came into contact during the incident. | Text (80) |
| | Highest Injury Severity | The Reporting Entity's report of the highest confirmed or alleged crash injury severity level resulting from the incident. | Text (80) |
| | Property Damage | The Reporting Entity's report of whether the incident resulted in any property damage. | Text (80) |
| | Any Air Bags Deployed | The Reporting Entity's report of whether any air bag in the Crash Partner / Other vehicle deployed during the incident. | Text (80) |
| | Incident Date | The Reporting Entity's report of the month and year the incident occurred [MMM-YY]. | Text (20) |
| | Incident Time | The Reporting Entity's report of the local time at which the incident | Text (10) |
| | | occurred [24-hour format]. | |
| | State | The state or territory where the incident occurred, as reported by the Reporting Entity. | Text (3) |
| | City | The city where the incident occurred, as reported by the Reporting Entity | TEXT(30) |
| | Latitude | The Reporting Entity's report of the latitude of the incident location in decimal degrees | Decimal (15) |
| | Longtitude | The Reporting Entity's report of the longtitude of the incident location in decimal degrees | Decimal (15) |
| Contact Area (ego & other) | Rear Left | whether the Reporting Entity reported any crash contact or damage resulting from the crash at the rear left on the crash partner or other vehicle. | Text ("Y", blank) |
| | Left | whether the Reporting Entity reported any crash contact or damage resulting from the crash at the left on the crash partner or other vehicle. | Text ("Y", blank) |
| | Front Left | whether the Reporting Entity reported any crash contact or damage resulting from the crash at the front left on the crash partner or other vehicle. | Text ("Y", blank) |
| | Rear | whether the Reporting Entity reported any crash contact or damage resulting from the crash at the rear on the crash partner or other vehicle. | Text ("Y", blank) |
| | Top | whether the Reporting Entity reported any crash contact or damage resulting from the crash at the top on the crash partner or other vehicle. | Text ("Y", blank) |
| | Bottom | whether the Reporting Entity reported any crash contact or damage resulting from the crash at the bottom of the crash partner or other vehicle. | Text ("Y", blank) |
| | Front | whether the Reporting Entity reported any crash contact or damage resulting from the crash at the front on the crash partner or other vehicle. A | Text ("Y", blank) |
| | Rear Right | whether the Reporting Entity reported any crash contact or damage resulting from the crash at the rear right on the crash partner or other vehicle. | Text ("Y", blank) |
| | Right | whether the Reporting Entity reported any crash contact or damage resulting from the crash at the right on the crash partner or other vehicle. | Text ("Y", blank) |
| | Front Right | whether the Reporting Entity reported any crash contact or damage resulting from the crash at the front right on the crash partner or other vehicle. | Text ("Y", blank) |